%% file: Cvpr.tex
\documentclass[final]{cvpr}
\usepackage{times}
\usepackage{epsfig}
\usepackage{graphicx}
\usepackage{amsmath}
\usepackage{amssymb}
\usepackage{adjustbox}
\usepackage{tabularx}
\usepackage[caption=false]{subfig}
\usepackage{color}
\usepackage{floatrow}
\usepackage[dvipsnames]{xcolor}
\usepackage{enumitem}
\input{latex/tools}
\usepackage{multirow}
\usepackage{colortbl}
\usepackage[pagebackref=true,breaklinks=true,colorlinks,bookmarks=false]{hyperref}

\def\cvprPaperID{5213} % *** Enter the CVPR Paper ID here
\def\confYear{CVPR 2021}

\begin{document}

%%%%%%%%% TITLE
%\title{First Person Audio-Visual Domain Generalization  \\ through 
%Relative Norm Alignment }
\title{Cross-Domain First Person Audio-Visual Action Recognition  \\ through 
Relative Norm Alignment }

%\author{First Author\\
%Institution1\\
%Institution1 address\\
%{\tt\small firstauthor@i1.org}

% For a paper whose authors are all at the same institution,
% omit the following lines up until the closing ``}''.
% Additional authors and addresses can be added with ``\and'',
% just like the second author.
% To save space, use either the email address or home page, not both
%\and
%Second Author\\
%Institution2\\
%First line of institution2 %address\\
%{\tt\small secondauthor@i2.org}
%}

\author{Mirco Planamente{\thanks{The authors equally contributed to this work. This paper is partially supported by the ERC project RoboExNovo. 
Computational resources were partially provided by IIT.}} $^{, 1,2}$ \quad
Chiara Plizzari\footnotemark[1] $^{, 1}$ \quad 
Emanuele Alberti\footnotemark[1] $^{, 1}$ \quad

Barbara Caputo\textsuperscript{1,2} \\

\and \textsuperscript{1} Politecnico di Torino\\
{\tt\small {name.surname}@polito.it}

\and \textsuperscript{2} Istituto Italiano di Tecnologia\\
{\tt\small {name.surname}@iit.it}
}

%\thanks{$^{*}$indicates equal contribution. 
%This paper is partially supported by the ERC project RoboExNovo. 
%Computational resources were partially provided by IIT.}    
%\thanks{$^{1}$ Politecnico di Torino, Corso Duca degli Abruzzi, 24 - 10129 Turin
%        {\tt\small {name.surname}@polito.it}}%
%\thanks{$^{2}$ Istituto Italiano di Tecnologia, Largo Rosanna Benzi, 10 - 16132 Genova  
%{\tt\small {name.surname}@iit.it}}%   

\maketitle

%%%%%%%%% ABSTRACT
\input{latex/Text/Abstract}

%%%%%%%%% BODY TEXT
\input{latex/Text/Intro}

\input{latex/Text/RelatedWorks}
\input{latex/Text/Method}

\input{latex/Text/Experiments}
\input{latex/Text/Conclusion}

{\small
\bibliographystyle{ieee_fullname}
\bibliography{egbib}
}

\end{document}

%% file: latex/tools.tex
\usepackage{xcolor} 
\usepackage{soul} 
\usepackage{xspace}
\usepackage{amsmath}

\newcommand{\mylossRna}{$\mathcal{L}_{RNA}$ }

%% file: latex/Text/Abstract.tex
\begin{abstract}
%Fine-grained first person action recognition has been widely studied by the community in the last years
%Fine-grained 
First person action recognition is an increasingly researched topic  because of the growing popularity of wearable cameras. This is bringing to light cross-domain issues that are yet to be addressed in this context. Indeed, the information extracted from learned representations suffers from an intrinsic environmental bias. This strongly affects the ability  to generalize to unseen scenarios, limiting the application of current methods in real settings where trimmed labeled data are not available during training. In this work, we propose to leverage over the intrinsic complementary nature of audio-visual signals to learn a representation  that works well on data seen during training, while being able to generalize across different domains. 
To this end, we introduce  an audio-visual loss that aligns the contributions from the two modalities by acting on the magnitude of their feature norm representations. This new loss, plugged into a minimal multi-modal action recognition architecture, leads to strong results in cross domain first person action recognition, as demonstrated by extensive experiments on the popular EPIC-Kitchens dataset.
%to a significant boost in performance across domains, both in Domain Generalization (DG) and Unsupervised Domain Adaptation Settings. Experiments on Epic-Kitchen, where we define the first DG protocol in first-person action recognition, show the power of our approach.   

%We provide a Domain Generalization benchmark for audio-visual data on the popular Epic-Kitchen dataset, achieving competitive results over the most recent Domain Adaptation methods. 

%videos taken from wearable devices suffer from egocentric motions and sudden changes of viewpoints. While these problems have been widely studied by the community, it is still uncovered the  

%while they do not affect auditive information
%raga quali sono i concetti che vorremmo esprimere nell'abstract?
%abbiamo tipo 3 punti chiave?

%DG 
%Audio

\end{abstract}

%% file: latex/Text/Intro.tex
\section{Introduction}
\begin{centering}

\begin{figure}[t]

    \includegraphics[width=1\linewidth]{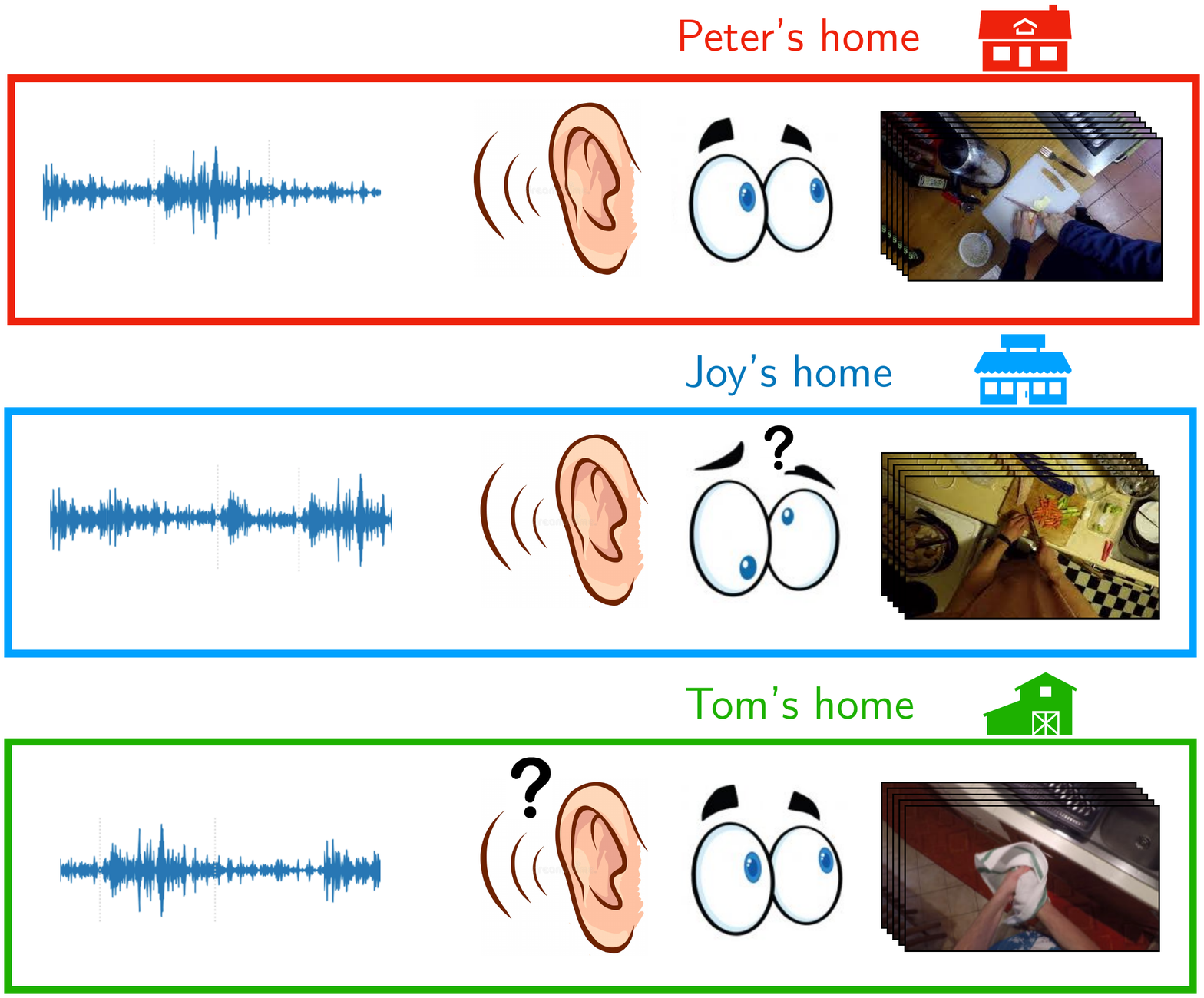}
    \caption{Egocentric action recognition comes with  a rich sound representation, due to the frequent hand-object interactions and the closeness of the sensor to the sound source. Here we show that the complementary nature of visual %(RGB)
    and audio information can be exploited to deal with the cross-domain challenge. %first person activity recognition. 
    %%the appearance information (RGB) and the audio signal,
    %proved to obtain good results in the context of first person action recognition. %Our insight is based on the assumption that both these two kinds of data
    %Our insight is that both modalities might suffer from the domain shift due to the inevitable environmental bias, %but we believe that the heterogeneous nature of them could be reflected in different ways in which the domain shift problems affect them.  
    %Our insight is that both modalities might suffer in different manners from the domain shift imputable to the environmental bias affecting each dataset. Hence, finding an efficient way to merge the two information to let one overcome the downsides of the other is crucial to resolve such issue.
%Hence, improving the cooperation between the two could help one modality to overcome the problem, introduced by the domain shift of the other, in a complementary way.
}
    \label{fig:OurLoss}
    
\end{figure}
\end{centering}

%Egocentric vision plays a central role in a wide range of applications, such as human-computer interaction or human assistance. The recent release of the EPIC-Kitchen large-scale dataset   is achieving more and more interest by the research community, especially 

First Person Action Recognition is rapidly attracting the interest of the research community \cite{sudhakaran2018attention,sudhakaran2019lsta,furnari2020rolling,Kazakos_2019_ICCV,ghadiyaram2019large,Wu_2019_CVPR}, both for the significant challenges it presents and for its central role in real-world egocentric vision applications, from wearable sport cameras to  human-robot interaction or human assistance. 
The recent release of the EPIC-Kitchen large-scale dataset \cite{damen2018scaling} has given a very significant boost to the research activities in this field, offering the possibility to study people’s daily actions from a unique point of view. The collection of this dataset consists in the segmentation of long untrimmed videos representing people’s daily activities recorded in the same kitchen. This process results in a huge number of sample clips representing a large variety of action classes, which are however captured in a limited number of environments. This intrinsic unbalance causes the so called environmental bias, meaning that the learned action representations are strictly dependent on the surroundings, and thus hardly able to generalize to videos recorded in different conditions \cite{torralba2011unbiased}. In general, this problem is referred to in the literature as domain shift, meaning that a model trained on a source labelled dataset cannot generalize well on unseen data, called target. Recently, \cite{Munro_2020_CVPR} addressed this issue %for fine-grained AR 
by reducing the problem to an unsupervised domain adaptation (UDA) setting, where an unlabeled set of trimmed samples from the target is available during training. However, the UDA setting is not always  realistic, because the target domain might not be known a priori or because it might be costly (or plainly impossible) to access target data at training time. 

In this paper we argue that the true challenge is to learn a representation %able to extract a knowledge from data which is transferrable 
able to generalize to any unseen domain, regardless of the possibility to access target data at training time. This means developing a method general enough to work both on UDA and Domain Generalization (DG) \cite{pmlr-v28-muandet13}. %, the extreme case where target data are not accessible at training time.
%The considerations above motivate us to analyze the problematics of this problem, known as domain generalization (DG),rained context. 
Inspired by the idea of exploiting the multi-modal nature of videos as done in \cite{Munro_2020_CVPR}, we propose a new cross-domain generalization method which leverages over the complementary nature of visual and audio information. We start by observing that first person action recognition intrinsically comes with rich sound information, due to the strong hand-object interactions and the closeness of the sensors to the sound source. The use of auditory information could be a good workaround for the problems which arise from the use of wearable devices, in that it is not sensitive to the ego-motion and it is not limited by the field of view of the camera.  Moreover, our idea is that, since the audio and visual modalities come from different sources, the domain-shift they suffer from is not of the same nature.  Motivated by these considerations, we propose a new cross-modal loss function, which we call Relative Norm Alignment loss, that operates on the relative features norm of the two modalities by acting on their magnitude. Our loss improves the cooperation between the audio and visual channels, which results in a stronger ability  to overcome the domain shift. We show with extensive experiments that, when used in a very simple audio-visual architecture, our loss leads to strong results  both in UDA and DG settings.  %of the other in a complementary way.

To  summarize,  our  contributions  are  the  following: 
\begin{itemize}
    \item  we empirically bring to light a problem related to the heterogenous nature of audio and visual modalities, which causes an unbalance preventing the two modalities to correctly cooperate; 
    \vspace{-7pt}
    \item we propose a new cross-modal audio-visual Relative Norm Alignment loss by progressively aligning the relative feature norms of the two modalities;
    \vspace{-7pt}
    \item we present a new benchmark for both single-source and multi-source DG settings in first person videos, which, to the best of our knowledge, no prior work has explored yet;
    \vspace{-7pt}
    \item we validate the effectiveness of our method on both DG and UDA scenarios, achieving competitive results compared to previous works.
    \vspace{-7pt}
\end{itemize}

%% file: latex/Text/RelatedWorks.tex
\section{Related Works}
\textbf{First Person Action Recognition.}
Until now, research has been focused on data provided by a specific view of the camera (often fixed), i.e., third person view \cite{10.5555/2968826.2968890,wang2016temporal,carreira2017quo}. With the recent release of a large-scale dataset of first-person actions \cite{damen2018scaling}, the community has also become interested in working on videos that are recorded from an egocentric point of view. 
Since egocentric action recognition suffers from the motion of the camera and sudden changes of view, the main approaches proposed so far are based on  multi-stream architectures \cite{carreira2017quo,10.5555/2968826.2968890,ma2016going,lin2019tsm,cartas2019seeing,Kazakos_2019_ICCV,lu2019learning}, many of which are inherited from the third-person action recognition literature. The networks used
to extract spatial-temporal information from egocentric videos can be divided into two main groups. The first exploits Long Short-Term Memory and variants \cite{Sudhakaran_2017_ICCV,sudhakaran2018attention,sudhakaran2019lsta,furnari2020rolling} to generate an embedding representation based on the temporal relations between the features frames. The second \cite{singh2016first,tran2015learning,Wu_2019_CVPR,kapidis2019multitask} leverages 3D convolutional kernels which jointly generate spatial-temporal features by sliding along the spatial and temporal dimensions. %generating jointly spatial-temporal features. 
Recent works exploit an attention mechanism at frame or clip level \cite{sudhakaran2018attention,sudhakaran2019lsta,perezrua2020knowing,Lu2019TIP,lu2019learning} to re-weight the spatial or temporal features, obtaining remarkable results.
By observing the importance of multi-stream approaches in this context, they \cite{sudhakaran2019hierarchical,wangsymbiotic,Wu_2019_CVPR,zhou2018temporal} investigate %in a more 
%accurate 
alternative methods to fuse streams w.r.t. the standard late fusion approach, creating a more compact multi-modal representation.  Although optical flow has proven to be a strong asset for the action recognition task, %because of its complementary nature with the appearance information and its invariance to changing environments, using it is %the cost that it required is 
it is computationally expensive. As shown in %the plot of \textit{accuracy vs. time} reported by 
\cite{Crasto_2019_CVPR}, %in figure 1, 
the use of optical flow limits the application of several methods in online scenarios, pushing the community either to investigate alternative paths \cite{Kazakos_2019_ICCV,cartas2019seeing} or towards single-stream architectures \cite{zhao2019dance,Crasto_2019_CVPR,lee2018motion,sun2018optical,planamente2020joint}.

\textbf{Audio-Visual Learning.}
A wide literature exploits the natural correlation between audio and visual signals to learn cross-modal representations that can be transferred well to a series of downstream tasks, such as third person activity recognition. Most of these representation learning methods use a self-supervised learning approach, consisting in training the network to solve a \textit{synchronization} task~\cite{look_listen_learn,multisensory_owens, cooperative_torresani,objects_that_sound,afourasself,aytar2016soundnet}, i.e., to predict whether the audio and visual signals are temporally aligned or not. By solving this pretext task, the network is induced to find a correspondence between audio and visual cues, making the resulting representations perfect for tasks like sound-source localization \cite{objects_that_sound,afourasself,Zhao_2018_ECCV}, active speaker detection \cite{out_of_time,afourasself}, and multi-speaker source separation \cite{multisensory_owens,afourasself}. Audio has also been used as a preview for video skimming, due to its lightweight characteristics \cite{listen_to_look}. More recently, it proved to be useful even in egocentric action recognition \cite{Kazakos_2019_ICCV,cartas2019seeing}. %Recently, audio revealed by Kazakos et al.  \cite{Kazakos_2019_ICCV} to be a robust information for Egocentric Action Recognition, showing promising results when .
However, the role of this information in a cross-domain context is still unexplored. In this work, we investigate the importance of audio when used together with visual information in learning a robust representation on unseen data.

\textbf{Unsupervised Domain Adaptation (UDA).}
The goal of UDA is to bridge the domain gap between a labeled source domain and an unlabeled target one.
%UDA is a well studied task \cite{survey-wang2018deep, survey-wilson2020survey, survey-kouw2019review} aiming to bridge the domain shift between a labeled source domain and an unlabeled target one.
We can divide UDA approaches in \textit{discrepancy-based} methods, which explicitly minimize a distance metric among source and target distributions \cite{da-afnxu2019larger, da-mcdsaito2018maximum}, e.g., the maximum mean discrepancy (MMD) in \cite{da-mmdlong2015learning}, and \textit{adversarial-based} methods~\cite{da-adv-deng2019cluster, da-adv-tang2020discriminative}, often leveraging a gradient reversal layer (GRL)~\cite{grl-pmlr-v37-ganin15}. %which reverses the gradient to make the generator produce domain-invariant representations to be passed to the task-specific classifier.
Other works exploit batch normalization layers to normalize source and target statistics \cite{ada-bn, DBLP:conf/iclr/LiWS0H17, da-bnchang2019domain}.
Still, another approach is the \textit{generative-based} one, which operates by performing style-transfer directly on input data \cite{da-cycle-gong2019dlow, da-cycle-hoffman2018cycada}.
The approaches described above have been designed for standard image classification tasks. Only few works analyzed UDA for video understanding \cite{videoda-chen2019temporal,Munro_2020_CVPR,videoda-choi2020unsupervised,videoda-Jamal2018DeepDA}. \cite{videoda-chen2019temporal} focuses on aligning temporal relation features to increase robustness across domains. In \cite{videoda-choi2020unsupervised}, the network is trained to solve an auxiliary self-supervised task on source and target data. Recently \cite{Munro_2020_CVPR} proposed an UDA method for first person fine-grained action recognition, called MM-SADA, combining a multi-modal self-supervised pretext task with an adversarial training. 

%The Adversarial approaches CITA and the Self-Supervised pretext tasks CITA are notice effective also in video context. 
%Very recently method [] provide a strategy to effectively align domains spatio-temporally for videos by aligning temporal relation features, achieving good performance in the context. 

\textbf{Domain Generalization (DG).} The DG setting is closer to real-world conditions, in that it addresses the problem of learning a model able to generalize well using inputs from multiple distributions, when no target data is available at all. Previous approaches in DG are mostly designed for image data \cite{carlucci2019domain,volpi2018generalizing,li2018domain,dou2019domain,li2018deep,bucci2020selfsupervised} and are divided in \textit{feature-based} and \textit{data-based} methods. The former focus on extracting invariant information which are shared across-domains~\cite{li2018domain,li2018deep}, while the latter exploits data-augmentation strategies to augment source data with adversarial samples and possibly get closer to the target~\cite{volpi2018generalizing}. Interestingly, using a self-supervised pretext task is an efficient solution to the extraction of a more robust data representation \cite{carlucci2019domain,bucci2020selfsupervised}. We are not aware of previous works on first or third person DG. 
%in video based context either for fine-grained action recognition and nor in First Person data, we find just
Among unpublished works, we found only one \textit{arXiv} paper~\cite{videodg-yao2019adversarial}, in third person action recognition, designed for single modality. Under this setting, first person action recognition models, and action recognition networks in general, degenerate in performance due to the strong divergence between source and target distributions. Our work stands in this DG framework, and proposes a feature-level solution to this problem in first person action recognition by leveraging the natural audio-visual correlation. %in this specific setting.

%% file: latex/Text/Method.tex
\section{Relative Norm Alignment}\label{sec:preliminaries}

%\subsection{Overview}
%Our goal is to learn a model representation which, using video clips of the source domain and their corresponding sound, is able to predict well over 
\subsection{Problem Statement}
Given one or more source domains $\{\mathcal{S}_1,...,\mathcal{S}_k\}$, where each $\mathcal{S}={\{(x^s_i,y^s_i)\}}^{N_s}_{i=1}$ is composed of $N_s$ source samples with label space $Y^s$ known, our goal is to learn a model representation able to perform well on a target domain $\mathcal{T}={\{x^t_i\}}^{N_t}_{i=1}$ of $N_t$ target samples whose label space $Y^t$ is unknown. Our two main assumptions are that the distributions of all the domains are different, \ie $\mathcal{D}_{s,i} \neq \mathcal{D}_t$ $\land$ $\mathcal{D}_{s,i} \neq \mathcal{D}_{s,j}$, with $i \neq j$, $i,j=1,...,k$, and that the label space is shared, $\mathcal{C}_{s,i} = \mathcal{C}_t$, $i=1,...,k$. In this work we consider two different scenarios:
%\begin{itemize}

\noindent
{\textbf{Domain Generalization (DG)}},
where at training time the model can access one or more fully labeled source datasets $\mathcal{S}_1,...,\mathcal{S}_m$, but no information is available about the target domain $\mathcal{T}$.

\noindent
{\textbf{Unsupervised Domain Adaptation (UDA)}}, where at training time it is possible to access a set of unlabeled target samples belonging to the target domain $\mathcal{T}$, jointly with one fully labeled source domain $\mathcal{S}$.

%Under this setting, during training time, in addition to one fully labeled source domain $\mathcal{S}$, a set of unlabeled target samples belonging to the target domain $\mathcal{T}$ is available. 

%During training time, the model disposes of one or more fully source labeled datasets $\mathcal{S}_1,...,\mathcal{S}_m$, but no information is available about the target domain  $\mathcal{T}$. Thus, the goal is to build a model able to generalize well on unseen domains. 
%\end{itemize}
\noindent
For both scenarios, the ultimate goal is to learn a classifier able to generalize well on the target data.

\textbf{Multi-Modal Approach.} Our goal is to investigate how using multi-modal signals from source and target data affects the ability of a first-person action classification net to generalize across domains.
%We are interested in a setting where the source and target data consist of multi-modal signals, with the aim to investigate how this affects the ability of a first-person action classification architecture to generalize across domains.
Specifically, given
a multi-modal input $X=(X^1,...,X^M)$, where $X^m=(x^m_1,...x^m_{N_m})$ is the set of all $N_m$ samples of the $m$-th modality, we use a separate feature extractor $F^m$ for  $X^m$%corresponding to the $m$-th modality
, and we employ all the $f_m=F^m(x^m_i)$ corresponding features, encoding information from multiple channels,  during the learning process. We denote with $h(x^m_i)=({\lVert{ \cdot }\rVert}_2 \circ f_m)(x^m_i)$ the $L_2$-norm of the features $f_m$. 

\subsection{Cross-Modal Audio-Visual Alignment}
\label{sec:assumptions}
\begin{figure}[t]
    \centering
    \includegraphics[width=1\linewidth]{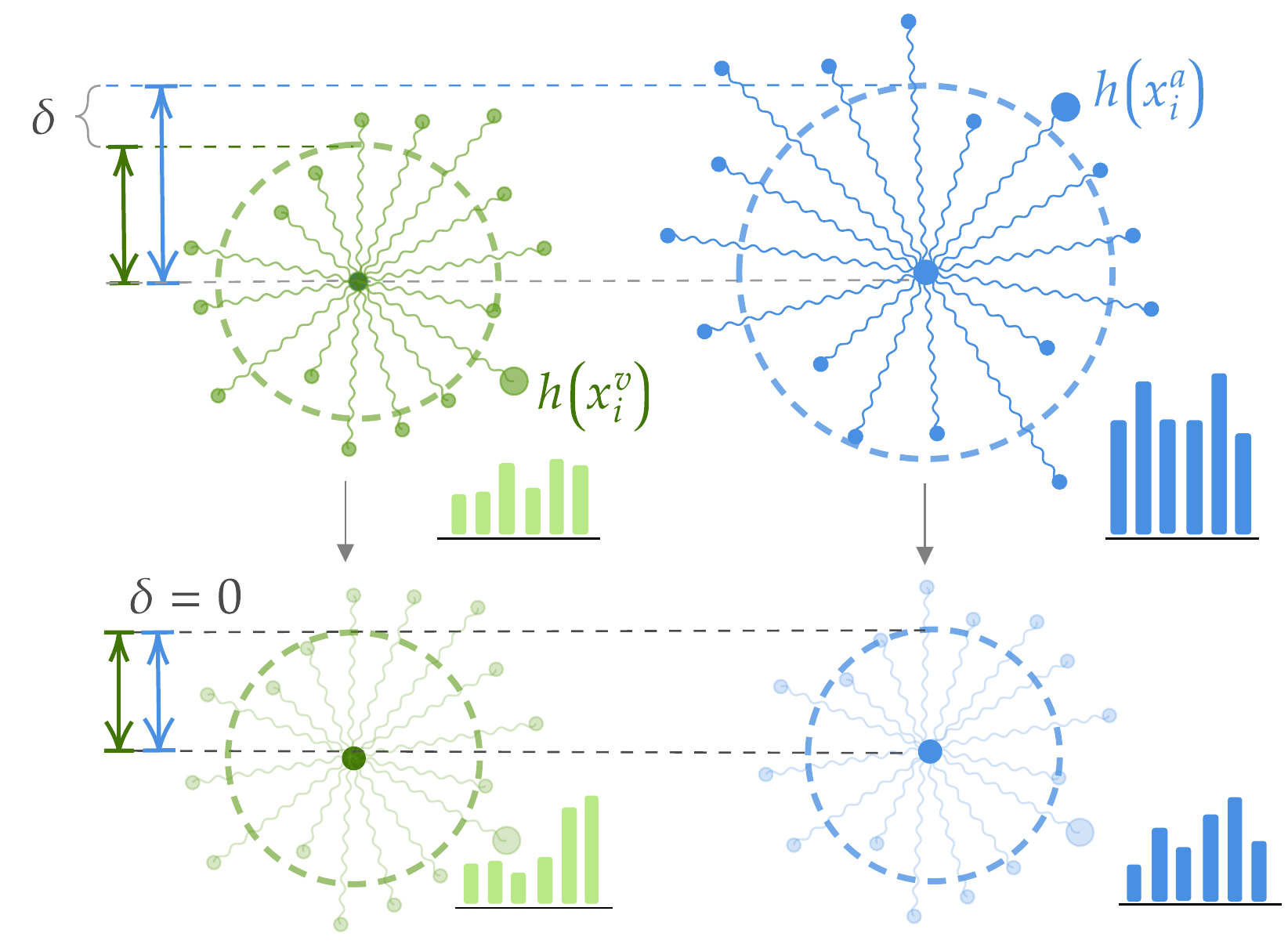}
    \caption{\textbf{Relative Norm Alignment.} The norm $h(x^v_i)$ of the $i$-th \textcolor{OliveGreen}{visual} sample (top-left) and $h(x^a_i)$ of the $i$-th \textcolor{RoyalBlue}{audio} sample (top-right) are represented, by means of segments of different lengths. The radius of the two circumferences represents the mean feature norm of the two modalities, and $\delta$ their discrepancy. 
    By minimizing $\delta$, audio and visual feature norms are induced to be the same, as shown at the bottom.
    %Through our $\mathcal{L}_{RNA}$, audio and visual features norms are aligned by inducing them to have the same value, i.e., same radius, while reducing the term $\delta$ to zero (Bottom).   }
    }
    \label{fig:Feat_Norm}
    \vspace{-0.2cm}
\end{figure}

Let us consider a multi-modal framework characterized by $M=2$ modalities, specifically RGB clips and audio signals. We indicate with $f_v=F^v(x^v_i)$ and $f_a=F^a(x^a_i)$ the features encoding the visual and audio information, respectively (details about the feature extractor modules are given in Section \ref{RNA-NET}).
The discrepancy between their norms, i.e., $ h(x^v_i)$ and $h(x^a_i)$, is measured by a \textit{mean-feature-norm distance} term $\delta$, defined as:
\begin{equation}
    \delta(h(x^v_i),h(x^a_i))=\frac{1}{N}\sum_{x^v_i \in \mathcal{X}^v}h(x^v_i)-\frac{1}{N}\sum_{x^a_i \in \mathcal{X}^a} h(x^a_i) ,
\end{equation} 
where $N=|\mathcal{X}^v|=|\mathcal{X}^a|$ denotes the number of the samples for each modality. Figure \ref{fig:Feat_Norm} illustrates the feature norms of the two modalities and the $\delta$ between the two.
%\barbara{Sarebbe importante qui fare piu' esplicitamente riferimento alla figura e a quello che rappresenta; parallelamente, forse si potrebbe aggiungere alla figura $\delta$? }

 It has been shown in the literature that aligning audio and visual information by solving synchronization tasks~\cite{look_listen_learn,multisensory_owens, cooperative_torresani,objects_that_sound,afourasself} leads to representations that facilitate a number of audio-visual downstream tasks, including action recognition. Such approaches enforce feature alignment by means of Euclidean or similarity-based losses, whose objective is to embed audio and visual inputs into a shared representation space.

%In literature, it has been shown that the alignment between audio and visual information \chiara{by solving a synchronization task} leads to good action recognition results \cite{look_listen_learn,multisensory_owens, cooperative_torresani,objects_that_sound,afourasself}.  by means of Euclidean and similarity-based losses, whose objective is to embed the audio and visual inputs into a shared space.

Our intuition is that the optimization of these loss functions could, to some extent, limit action recognition networks when dealing with cross-modal scenarios.
%in exploiting the cross-modal potentiality. 
This is because, as opposed to acting on the magnitude of audio and visual norms, these losses mainly use the angular distance $\theta$ between the two embeddings, defined as $\theta=arccos(\frac{\mathbf{f}_v\cdot \mathbf{f}_a}{\lVert{f_v}\rVert\lVert{f_a}\rVert})$. 
By acting only on the normalized feature vectors, they are indeed capable of aligning the two representations (Figure \ref{fig:OurLoss}-b) but they struggle to exploit the modality-specific characteristics of the two streams. 
%As such, they induce a common representation, but limits the potential of the modality-specific ones, forcing an overlapping between two information which are not necessarily coherent. 
In other words, when using an angular loss we impose %on the network %is induced with 
the prior that what is significant for the visual stream is significant also for the audio stream, but this might not be true in practice, especially when training and test data come from different distributions. 
For instance, in a clip where the action \textit{take} does not produce any sound, %because of the object it involves, or the setting where it is performed, 
the information brought by the audio will be referred to only as background noise. Conversely, for the same action carried out with another object or in a different setting, the aural information might be highly informative, possibly even more than the visual one. 
%\barbara{ho modificato l'ultima frase per renderla specifica del cross-modal, please check!} 
We show below with a set of experiments (Section \ref{sec:experimental}, Figure \ref{fig:PlotAccNorm}) that a large $\delta$, i.e., a misalignment at feature-norm level, negatively affects the learning process by causing an unbalance between the contributions brought by each modality, which therefore degrades the classification performance.

%we say $f(a_i)$ is \textit{orthogonal} to $f(v_i)$, i.e., $f(a_i)$ $\perp$ $f(v_i)$, if their modality-specific information is complementary. Instead, we say $f(a_i)$ is \textit{parallel} to $f(v_i)$, i.e., $f(a_i)$ $\parallel$ $f(v_i)$, if the information is shared between the two. 

%We define $g(v_i)=({\lVert{ \cdot }\rVert}_2 \circ f(a_i))(v_i)$ and $g(a_i)=({\lVert{ \cdot }\rVert}_2 \circ f(a_i))(a_i)$ as the $L_2$-norms of the features extracted from the visual and audio streams respectively. We characterize their distance by defining the mean-feature-norm discrepancy between them, defined as:

%\begin{equation}
%    \delta(g(v_i),g(v_i))=\frac{1}{n_v}\sum_{v_i \in \mathcal{M}_v}g(v_i)-\frac{1}{n_a}\sum_{a_i \in \mathcal{M}_a} g(a_i) 
%\end{equation}
%We observe that a large $\delta$ corresponds to a misalignment between the two modalities which is revealed as one modality drowning out the other, preventing the network to exploit their orthogonality or parallelism. 

%This has as a side-effect a strong unbalance between the final predictions of the two, preventing them to cooperate to the final classification.

\subsection{Relative Norm Alignment Loss} \label{sec:rna_loss}
Motivated by the  considerations above, we propose a new cross-modal loss function, which aims to reduce the discrepancy between feature distributions by aligning their contribution during training. As opposed to losses acting on the normalized feature vectors, our loss operates on their magnitude, which intuitively results in more freedom to preserve the modality-specific features (see Figures \ref{fig:OurLoss}-b-c). We expect this to be important in cross-domain scenarios. Considering the dot product $<f_v, f_a>$, defined as
\begin{equation}
\label{formula:dot_product}
    <f_v, f_a>= {\lVert f_v \rVert}_2 {\lVert f_a \rVert}_2 \cos{\theta} ,
\end{equation}
our approach involves the first two terms of Equation \ref{formula:dot_product}, imposing a relative alignment between them. \\ Our relative norm alignment (RNA) loss function is defined as
\begin{equation}\label{formula:rna}
    \mathcal{L}_{RNA}=\frac{1}{N}\frac{\sum_{x^v_i \in \mathcal{X}^v}h(x^v_i)}{\sum_{x^a_i \in \mathcal{X}^a}h(x^a_i)} \rightarrow 1 ,
\end{equation}
where $h(x^v)=\lVert{ f_v }\rVert_2$ and $h(x^a)=\lVert{ f_a }\rVert_2$ indicate the norm of visual and audio features respectively. This dividend/divisor structure is used to encourage a relative adjustment between the norm of the two modalities, enhancing an \textit{optimal equilibrium} between the two embeddings. When minimizing \mylossRna, the network can either increase the divisor $(f_v \nearrow)$ or decrease the dividend $(f_a \searrow)$, leading to three potential main benefits:
\begin{figure}[t]
    \centering
    \includegraphics[width=1\linewidth]{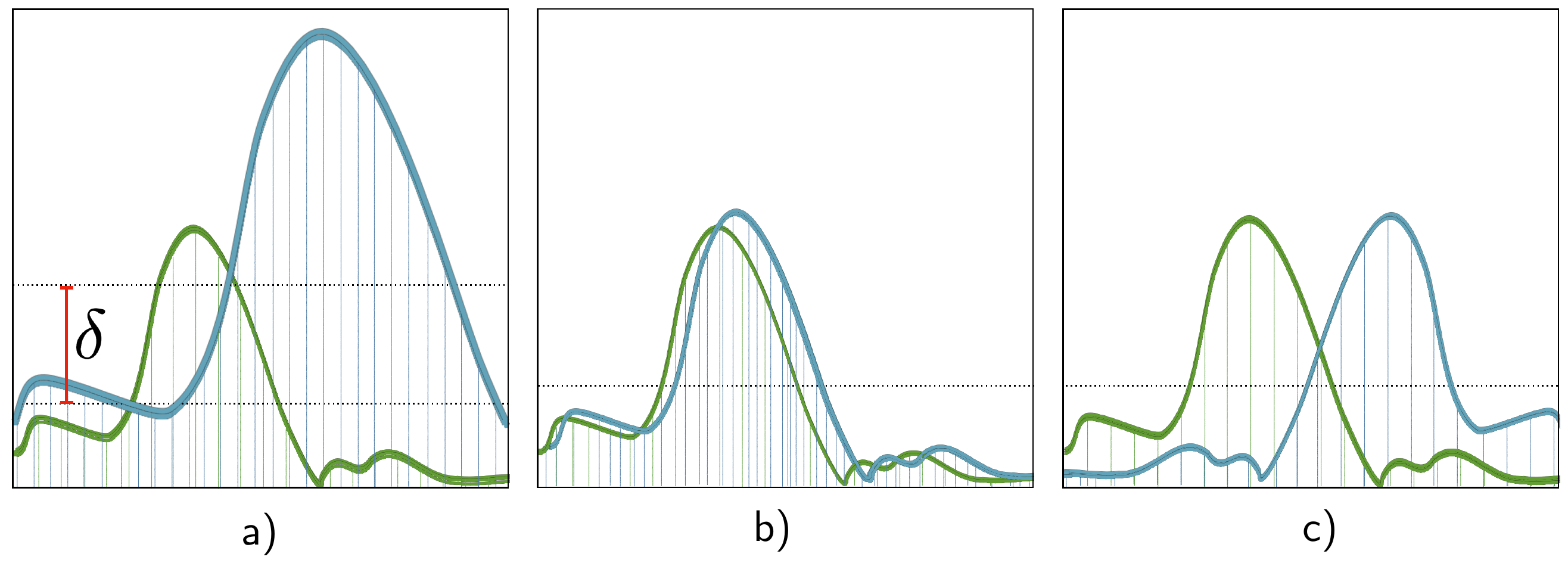}
    \caption{\textbf{Feature Distribution Adjustment.} (a) shows the distribution of \textcolor{OliveGreen}{visual} and \textcolor{RoyalBlue}{audio} features. We see that, without any form of alignment, audio features are predominant over visual ones, which could ultimately lead to a loss of information. By minimizing $\mathcal{L}_{RNA}$, two possible scenarios can occur, displayed in (b) and (c). In both, the range where the feature norms vary is the same, making the informative content of the two distributions comparable with each other. This lets the loss learn from data when it is more convenient to align them (b) or when it is better to preserve their peculiarities (c).
    }
    \label{fig:OurLoss}
    \vspace{-0.2cm}
\end{figure}
\begin{enumerate}[itemsep=1pt, topsep=0pt, partopsep=0pt]
    \item Since $\lVert{ f_v }\rVert_2$ and $\lVert{ f_a }\rVert_2$ tend to the same value, features norm ranges are encouraged to be comparable, preventing one modality to drown out the other, improving the final prediction (Figures \ref{fig:OurLoss}-b and \ref{fig:OurLoss}-c).
    \item By reducing the norm of the features while learning the feature extractor, the latter is free to choose which are the less/more discriminative ones, and lower/rise their norm accordingly, increasing the generalization ability of the model.
    \item Comparing to standard similarity losses, by not constraining the angular distance $\theta$ between the two modality representations, feature distributions have the freedom to arrange in non-overlapping configurations (Figure ~\ref{fig:OurLoss}-c).
\end{enumerate}
%\textbf{1)} Since $\lVert{ f_v }\rVert_2$ and $\lVert{ f_a }\rVert_2$ tend to the same value, features norm ranges are encouraged to be comparable, preventing one modality drowning out the other and improving the final classification prediction.
%\textbf{2)} By reducing the norm of the features while learning the feature extractor, the latter is free to choose which are the less/more discriminative information, and lower/rise their norm accordingly, increasing the generalization ability of the model.
%\textbf{3)} Comparing to standard similarity losses, by not constraining the angular distance $\theta$ between the two modality representations, features distributions have the possibility to arrange in non-overlapping configurations, exploiting the complementarity of the two modalities.
The effects observable at feature-level of the application of our \mylossRna are represented in Figure \ref{fig:OurLoss}. %for better comprehension. 
In Figure \ref{fig:OurLoss}-a we represent the features distribution learned with standard cross-entropy loss. As it can be seen, the feature norms of the two modalities differ by a $\delta$, meaning that the respective features lie within different ranges which make the two representations hard to compare. The solution proposed by our \mylossRna  corresponds to \textbf{1)}. In Figures \ref{fig:OurLoss}-b and \ref{fig:OurLoss}-c we show the feature representations obtained by minimizing our loss function. The situation depicted in Figure \ref{fig:OurLoss}-b occurs when audio and visual information ``agree" by means of their modality-specific features. %This recalls the ``ideal" alignment condition aimed for by standard similarity losses. 
The scenario depicted in Figure \ref{fig:OurLoss}-c is the most interesting, since it represents a situation which is not compatible with the aim of standard similarity losses. As stated in \textbf{3)}, our \mylossRna ensures that the modality-specific features are preserved, allowing the final classifier to exploit their complementarity.  

\vspace{-0.2cm}
\section{ Cross-Domain Audio-Visual RNA-Net} \label{RNA-NET}
\begin{figure}[t]
    \centering
    \includegraphics[width=\linewidth]{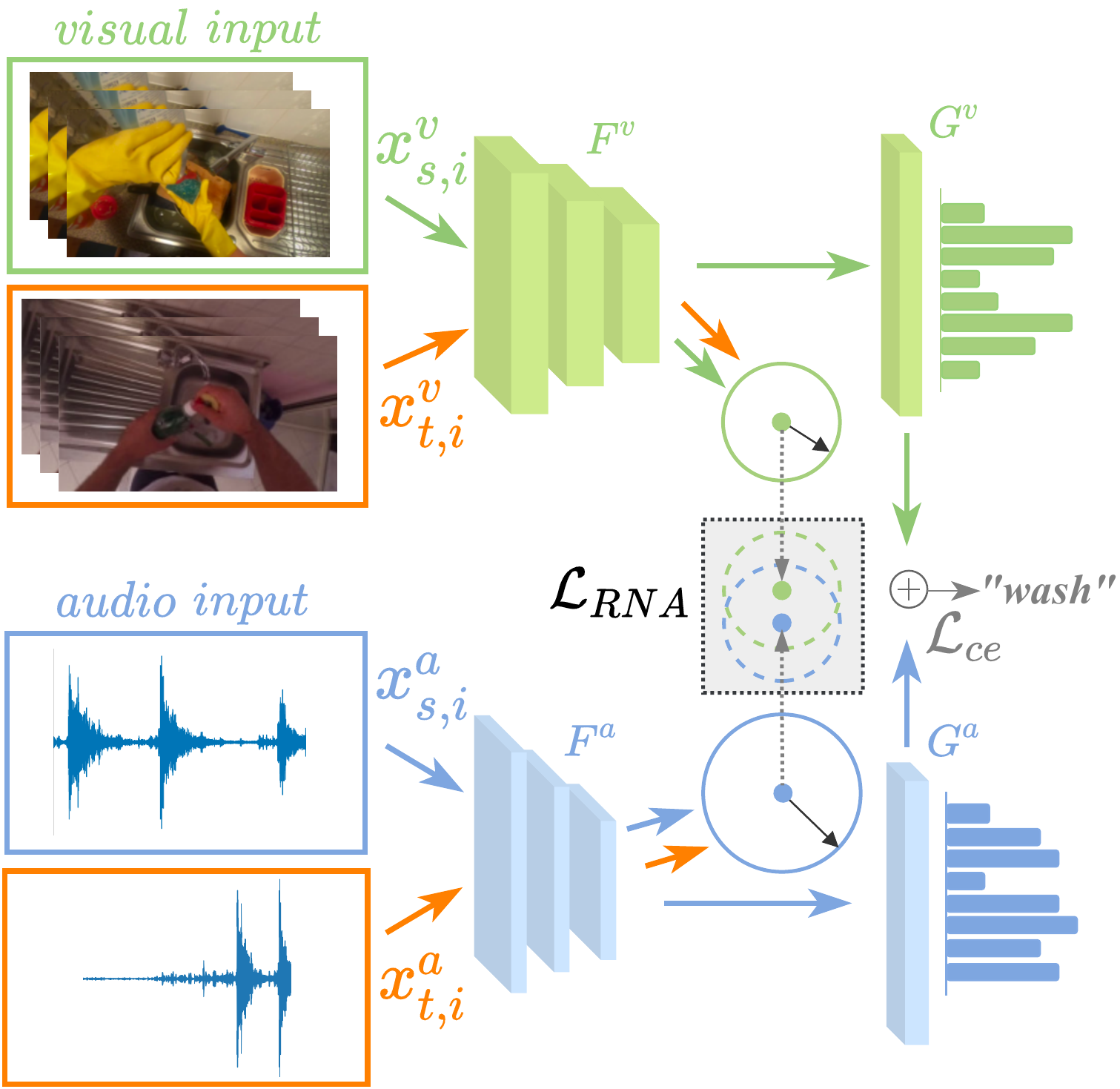}
    \caption{\textbf{RNA-Net.} Labeled \textcolor{OliveGreen}{source visual} $x^v_{s,i}$ and \textcolor{RoyalBlue}{source audio} $x^a_{s,i}$ inputs are fed to the respective feature extractors $F^v$ and $F^a$.  \textcolor{Orange}{Unlabeled target} data of any modality ($x^m_{t,i}$) is seen at training time only in UDA setting, and not in DG. Our $\mathcal{L}_{RNA}$ operates at feature-level by balancing the relative feature norms of the two modalities. The action classifiers $G^v$ and $G^a$ are trained with standard cross-entropy loss $\mathcal{L}_{ce}$. At inference time, multi-modal target data is used for classification.
    }
    
    %\textbf{RNA-Net.} \textcolor{OliveGreen}{Source} visual $x^v_{s,i}$ and \textcolor{Orange}{target} visual inputs $x^v_{t,i}$ are fed to the visual stream. Conversely, the \textcolor{RoyalBlue}{source} audio input $x^a_{s,i}$ and \textcolor{Orange}{target} audio input $x^a_{t,i}$, are fed to the audio stream. Feature extractors are separate for each modality, and shared for both source and target data. Our $\mathcal{L}_{RNA}$ operates at feature-level by balancing the relative feature norms of the two modalities. The audio and visual classifiers are trained with standard cross-entropy $\mathcal{L}_{entropy}$ using source data. During inference, multi-modal target data are used for classification. Note that in the DG scenario target data is not available during training.   }
    
    \label{fig:architecture}
    \vspace{-5 pt}
\end{figure}

\vspace{-0.05cm}
This section shows how \mylossRna %Relative Norm Alignment (RNA) 
 can be effectively used in a very simple audio-visual deep network for cross-domain first person action recognition. The network, shown in Figure \ref{fig:architecture}, inputs audio and visual information in two separate branches. After a separate convolution-based feature extraction step, the \mylossRna loss learns how to combine the two modalities, leading to a significant generalization across-domains, in both the domain generalization (DG) and unsupervised domain adaptation (UDA) settings. Below we describe in more details how the net works in both settings.

\subsection{AV-RNA-Net for Domain Generalization}
\label{sec:dg}

In a cross-modal context, the input comes from one or more source domains $\mathcal{S}_k=(\mathcal{S}^v,\mathcal{S}^a)$. Under the DG setting, the target is not available during training (see Section~\ref{sec:preliminaries}). % For simplicity, we denote with $X^m=(\mathcal{S}_m, \mathcal{T}_m)$ the set of inputs for the $m$-th modality, where $x^m=x^m_s$ while training, while $x^m=(x^m_s,x^m_t)$ is a tuple composed by both a source and target sample at inference time. 
As shown in Figure \ref{fig:architecture}, each input modality is fed to a separate feature extractor, $F^v$ and $F^a$ respectively. The resulting features $f_v=F^v(x^v_i)$ and $f_a=F^a(x^a_i)$ are then passed to separate classifiers $G^v$ and $G^a$, whose outputs correspond to distinct score predictions (one for each modality). The two are combined with a \textit{late fusion} approach and used to obtain a final prediction $P(x)$ (please refer to Section \ref{sec:experimental} for more details). Our loss \mylossRna operates at feature-level before the final classification, acting as a bridge between the two modalities and encouraging a balance between the respective contributions.
%Following previous works on multi-modal action recognition \cite{Munro_2020_CVPR, quo_vadis, tsn, trr, tsm}, we use a multi-stream structure, i.e., a separate stream for the visual and audio input, a
\par{\textbf{Why should \mylossRna help to generalize?}} Our loss \mylossRna rises the generalization ability of the network for two main reasons. \textbf{1)} By self-reweighting the two modalities contribution during training, the classifier has the chance to rank such contributions according to their real relevance, thus avoiding to be fooled by the natural unbalance due to their intrinsic heterogeneity. This is helpful especially in a multi-source setting, as it ensures uniformity not only across modalities, but also across data from different sources. \textbf{2)} By undirectly minimizing the norm of the feature activations, our loss sets a limit to the learner feature encoding, and thus it forces it to only learn the discriminative information. As a consequence, the classifier can learn to ignore information which is strictly domain-specific, distilling the most useful and transferable features.

% Given a visual input $X^v=(x^v_1,...x^v_N)$ and an audio input $X^a=(x^a_1,...x^a_N)$, $x^v_i \in \mathbb{R}^{T\times H\times W}$ denotes the $i$-th visual sample corresponding to a video clip, and $x^a_i \in \mathbb{R}^{T\times H\times W}$ denotes the $i$-th audio sample, represented as a spectrogram image.

\subsection{Extension to Unsupervised Domain Adaptation}\label{dg}
Under this setting, both labelled source data from a single source domain $\mathcal{S}=(\mathcal{S}^v,\mathcal{S}^a)$, and unlabelled target data $\mathcal{T}=(\mathcal{T}^v$, $\mathcal{T}^a)$ are available during training. Figure \ref{fig:architecture} shows the flow of source and target data, indicating with different colors source visual data (green), source audio data (blue) and target data (orange). We denote with $x_{s,i}=(x^v_{s,i},x^a_{s,i})$ and $x_{t,i}=(x^v_{t,i},x^a_{t,i})$ the $i$-th source and target samples respectively.  As it can be seen from Figure \ref{fig:architecture}, both $x^m_{s,i}$ and $x^m_{t,i}$ are fed to the feature extractor $F^m$ of the $m$-th specific modality, shared between source and target, obtaining respectively the features $f_s=(f^v_s,f^a_s)$ and $f_t=(f^v_t,f^a_t)$. In order to consider the contribution of both source and target data during training, we redefine our \mylossRna under the UDA setting as %i as shown in Equations \ref{eq:loss_s_t}, \ref{eq:loss_s_t_1} and \ref{eq:loss_s_t_2}.
\begin{equation}\label{eq:loss_s_t}
    \mathcal{L}_{RNA}=\mathcal{L}^s_{RNA}+\mathcal{L}^t_{RNA} ,
\end{equation}
\begin{equation}\label{eq:loss_s_t_1}
    \mathcal{L}^s_{RNA}=\frac{1}{N}\frac{\sum_{x^v_{s,i} \in \mathcal{X}^v_s}h(x^v_{s,i})}{\sum_{x^a_{s,i} \in \mathcal{X}^a_s}h(x^a_{s,i})} \rightarrow 1 ,
\end{equation}
\begin{equation}\label{eq:loss_s_t_2}
    \mathcal{L}^t_{RNA}=\frac{1}{N}\frac{\sum_{x^v_{t,i} \in \mathcal{X}^v_t}h(x^v_{t,i})}{\sum_{x^a_{t,i} \in \mathcal{X}^a_t}h(x^a_{t,i})} \rightarrow 1 ,
\end{equation}
By minimizing $\mathcal{L}^s_{RNA}$ we benefit from the considerations described in Section \ref{sec:dg}. Also, by minimizing $\mathcal{L}^t_{RNA}$, and thus learning the reweighting between the modalities on the unlabelled data, the encoded features contain useful information which directly enable us to adapt to the target. 

A problem that often occurs with UDA methods is that forcing an alignment between source and target features increases the risk of affecting the discriminative charateristics of the two, and thus destroying the inter-class separability~\cite{pmlr-v97-liu19b}. 
In our UDA setting, by operating on the two domains through separate $\mathcal{L}^s_{RNA}$ and $\mathcal{L}^t_{RNA}$, we mitigate this risk, preserving the discriminative structure of the two.

%% file: latex/Text/Experiments.tex
\section{Experiments}
\label{sec:experimental}
%\begin{itemize}
 %   \item Introduction
%\end{itemize}
%\mirco{da rivedere}
\subsection{Experimental Setting}\label{experimental_setting}
\noindent
\textbf{Dataset.} 
%In order to validate our approach in both Domain Generalization and Domain Adaptation settings (contexts) for fine-grained action recognition, 
Using the EPIC-Kitchens-55 dataset \cite{damen2018scaling}, we adopted the same experimental protocol of \cite{Munro_2020_CVPR}, where the three kitchens with %(P08, P01 and P22) having 
the largest amount of labeled samples are handpicked from the 32 available. We refer to them here as D1, D2, and D3 respectively. Since the action classification task is complicated by the large number of action labels, we consider only a subset, namely: \textit{put}, \textit{take}, \textit{open}, \textit{close}, \textit{wash}, \textit{cut}, \textit{mix}, and \textit{pour}. The challenges are not only due to the large domain shift %intrinsically existing 
among different kitchens, but also to the unbalance of the class distribution intra- and inter-domain, as shown in \cite{Munro_2020_CVPR}.

\noindent
\textbf{Input.}
Regarding the RGB input, a set of 16 frames, referred to as \textit{segment}, is randomly sampled during training, while at test time 5 equidistant segments spanning across all clips are fed to the network. At training time, we apply random crops, scale jitters and horizontal flips for data augmentation, while at test time only center crops are applied. 
Regarding aural information, we follow
%has to treated as an image in order to be processed by a CNN, similarly to what has been done in literature 
\cite{Kazakos_2019_ICCV} and convert the audio track into a 256 $\times$ 256 matrix representing the log-spectrogram of the signal. The audio clip is first extracted from the video, sampled at 24kHz and then the Short-Time Fourier Transform (STFT) is calculated of a window length of 10ms, hop size of 5ms and 256 frequency bands.% Hence, the x and y axis represent time and frequency, respectively.

%\textbf{Architecture.}
\noindent
\textbf{Implementation Details.} Our network is composed of two streams, one for each modality $m$, with distinct feature extractor $F^{m}$ and classifier $G^{m}$. The RGB stream uses the Inflated 3D ConvNet (I3D) with the same initialization proposed by the authors \cite{carreira2017quo},
%(pretrained on Kinetics \cite{kinetics} with a network initialized with ImageNet \cite{imageNet} pretrained weights) 
%\barbara{la parte tra parentesi e' un livello di dettaglio che si puo' spostare nei supplementary},
as done in
%in previous works in first person action recognition
\cite{Munro_2020_CVPR}. %to be coherent to previous action recognition methods both in third \cite{} and first person \cite{Munro_2020_CVPR}.
The audio feature extractor uses the BN-Inception model \cite{bn-inception}, a 2D ConvNet pretrained on ImageNet \cite{imageNet}, which proved to be a reliable backbone for the processing of audio spectrograms~\cite{Kazakos_2019_ICCV}. Each $F^{m}$ produces a 1024-dimensional representation $f_{m}$ which is fed to the action classifier $G^{m}$, consisting in a fully-connected layer that outputs the score logits for the 8 classes. Then, the two modalities are fused by summing the outputs and the cross entropy loss is used to train the network. %The domain classifier $D^{m}$, when present, is composed of two fully-connected layers, where the hidden layer is 100-dimensional, and a gradient reversal layer (GRL) \cite{grl-pmlr-v37-ganin15} to invert the gradient when backpropagating. \emanuele{ho omesso dettagli legati avgpool e cose di questo genere. se necessari li aggiungiamo}
To remain coherent with the setup used by \cite{Munro_2020_CVPR},
%and for fairer comparison, 
%most hyper-parameters are equal. 
we follow their strategy to validate our hyper-parameters.
All training models are run for 9000 iterations and finally tested with the average of the last 9 models. For further details on the optimizer, learning rate, parameters used and on the training process in general, we refer to the supplementary material. 
%\barbara{da qua in poi fino al commento di Ema sposterei tutto nei supplementary} 
%All experiments are run with batch size of 128, evenly split between source and target domains. The optimizer is Stochastic Gradient Descent (SGD) with momentum 0.9 and a weight decay of $10^{-7}$ and $4 \cdot 10^{-5}$ for RGB and audio streams, respectively. When not specified otherwise, the learning rate is 0.01 for both streams, divided by 10 at iter 3k for RGB and at iters 1k, 2k, and 3k for audio. \emanuele{mancano dettagli su domain classifier}
%To reproduce AdaBN \cite{ada-bn} each batch normalization layer is split into two to separately calculate statistics for source and target samples and use only the target ones when testing. The RGB batch normalization layers have momentum 0.9 and epsilon 0.001.
%\barbara{anche questo ultimo paraagrafo lo sposterei nei supplementary, tra l'altro discutiamo di come e' stato riprodotto AdaBN senza aver neanche detto che lo usiamo come baseline..}
\begin{figure}

\centering
    \includegraphics[width=\linewidth]{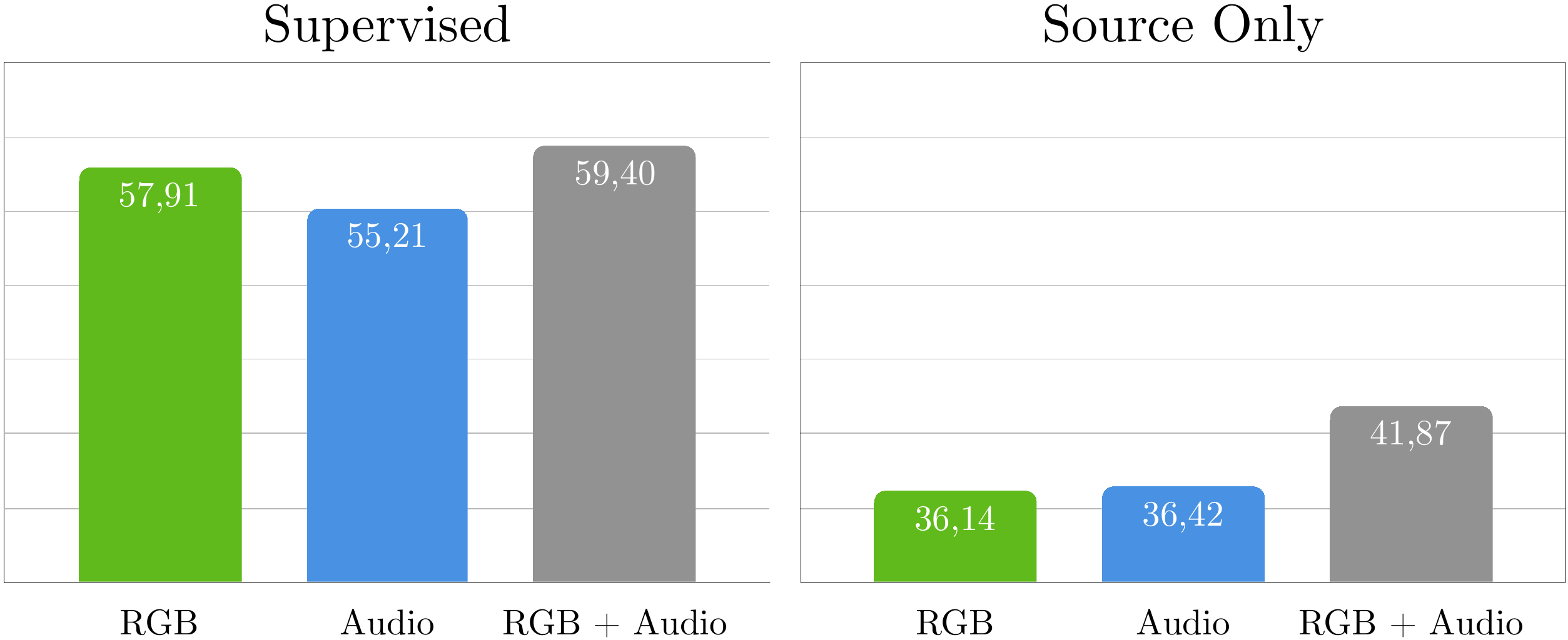}
    \caption{Single- vs multi-modality accuracy ($\%$) on both supervised and source only settings. The drop in performances when testing on target (right) highlight the presence of a strong domain shift.}
\label{fig:supervised_so}
\vspace{-0.2cm}
\end{figure}

\subsection{Results}
%%% ------------------- DG tables --------------------
\begin{table*}[ht]
\begin{minipage}{0.4\linewidth}

\begin{adjustbox}{width=1.4\columnwidth, margin=0ex 1ex 0ex 0ex}
\begin{tabular}{l|cccccc|c}
% OPPURE
%\begin{tabularx}{\columnwidth}{X|XXXXXXX}

\hline \hline

Single-Source & \multicolumn{1}{c}{D1 $\rightarrow$ D2} & D1 $\rightarrow$ D3 & D2  $\rightarrow$ D1     & D2 $\rightarrow$  D3     & D3  $\rightarrow$ D1     & D3  $\rightarrow$ D2   & Mean   \\ \hline \hline
Source Only &    39.03&	39.17	&35.27&	47.52&	40.26&	49.98&	41.87  \\ \hline 
Align. Only   &    38.50                        &   33.75    &     32.59       &    45.78       &    39.97       &      50.86 & 41.76     \\ %\hline
Orth. Only     &    39.18                        &    37.55    &      36.86      &     47.09      &      43.70    &       51.61   & 42.67 \\ %\hline
BatchNorm    &   40.03                        &    39.88    &      36.39      &    48.47     &     \underline{42.60}     &      48.33    & 42.62 \\ %\hline
SS \cite{Munro_2020_CVPR}  &    38.86                        &    33.75   &      32.59      &     45.78     &       39.97     &       50.86    & 40.30 \\ \hline

RNA-Net (Ours) &      \underline{45.01}                      &  \underline{44.62}   &    \underline{41.76}     &    \underline{48.90}       &       42.20  &    \underline{51.98} & \textbf{45.75}       \\ 

\hline

%\end{tabularx}

\end{tabular}

\end{adjustbox}

\end{minipage}
\hfill
\begin{minipage}{0.421\linewidth}

\begin{adjustbox}{width=1\columnwidth, margin=0ex 1ex 0ex 0ex}
\begin{tabular}{l|ccc|c}
\hline \hline
Multi-Source  & \multicolumn{1}{c}{D1, D2  $\rightarrow$ D3} & D1, D3  $\rightarrow$ D2 & D2, D3  $\rightarrow$ D1 & Mean\\ \hline \hline
Deep All      &       51.47                    &   43.19   & 39.35 & 44.67 \\ \hline
Align. Only       &       50.01                   &   42.40   & 44.40 & 45.60 \\ %\hline
Orth. Only     &       53.08                  &    41.76  &  48.07 &  47.64\\ %\hline
BatchNorm     &       52.07                    &   42.63  &  45.14 &  46.61 \\ %\hline
SS  \cite{Munro_2020_CVPR}          &       51.87                     &  39.79   & \underline{52.73} & 48.13    \\ \hline
RNA-Net (Ours) &      \underline{55.88}                     &   \underline{45.65}    &   51.64 &  \textbf{51.06} \\ 
\hline
\end{tabular}
\end{adjustbox}

\end{minipage}
\caption{Top-1 Accuracy ($\%$) of our RNA-Net under the single-source DG setting (left) and the multi-source DG setting (right).}
\label{tab:single-source-dg}
\vspace{-0.2cm}
\end{table*}

%%% ------------------- UDA table --------------------
\begin{table*}[ht]

\centering
\begin{adjustbox}{width=0.7\columnwidth, margin=0ex 1ex 0ex 0ex}
\begin{tabular}{l|cccccc|c}
\hline \hline
UDA              & \multicolumn{1}{c}{D1 $\rightarrow$ D2} & D1 $\rightarrow$ D3 & D2  $\rightarrow$ D1 & D2 $\rightarrow$ D3 & D3  $\rightarrow$ D1 & D3  $\rightarrow$ D2    & Mean \\ \hline \hline
SS Only \cite{Munro_2020_CVPR}         &       44.83                   &  42.88     &   40.61   &   \underline{54.21}   &   42.58    &    53.50 & 46.44  \\ %\hline
Adversarial Only \cite{grl-pmlr-v37-ganin15}        &          41.02                  &   43.04     &  39.36     &  49.25      & 38.77      &     50.56     &      43.67     \\ %\hline
MM-SADA \cite{Munro_2020_CVPR}         &         \underline{48.90}                   &  46.66    &  39.51    &   50.89    & \underline{45.42}      &\underline{55.14} & 47.75        \\ %\hline
MMD \cite{da-mmdlong2015learning}         &        42.40                   &  43.84     &  40.87    &  48.13   & 41.46        &50.03  & 44.46        \\ %\hline
AdaBN \cite{ada-bn}         &       36.64                   &  42.57    & 33.97      &   46.63    & 40.51        &  51.20 & 41.92        \\ \hline

RNA-Net (Ours)      &       46.89                  &  48.40   &   41.58   &  51.77    &   43.19    &    54.43 & 47.71  \\ %\hline
RNA-Net+GRL (Ours)      &       46.65                 &  \underline{49.95}   &  \underline{46.06}   &  51.77    &   42.20    &    53.14 & \textbf{48.30 } \\ \hline

\end{tabular}
\end{adjustbox}
\caption{Top-1 accuracy ($\%$) of our RNA-Net under the multi-modal UDA setting.}
\label{tab:multi-modal-da}
\vspace{-0.3cm}

\end{table*}
\noindent
\textbf{Preliminary Analysis.} To verify that combining audio and visual modalities actually improves results, we assess the impact of each modality individually (Figure \ref{fig:supervised_so}). Firstly, the two streams are trained both separately and jointly in a supervised fashion (referred to as \textit{supervised}). Then, we validate the same models under a cross-domain setting, meaning that training is performed on source data only, and test on unseen target data (referred to as \textit{source-only}).
%, in order to evaluate the extent of the domain shift.

Results (Figure \ref{fig:supervised_so}, left) highlight that, by using a single domain, the visual part is more robust than the audio one (+$2.7\%$). Conversely, when testing on target data from a different domain (Figure~\ref{fig:supervised_so}, right), audio is on-par with RGB. %audio-only achieves performances on-par with the rgb-only part.
This suggests that when a domain-shift exists, it is mainly imputable to changes in visual appearance. In the cross-domain scenario, %it has to be noticed that 
the accuracy drops dramatically (Figure~\ref{fig:supervised_so}, right), proving how the domain shift impacts negatively the performance. Interestingly, we see that the fusion of the two modalities brings a greater contribution when facing this problem, increasing the source-only results on single modality by $4\%$. This confirms that combining audio and visual cues is useful to partially overcome the weaknesses of each individual modality across domains. A similar exploration on audio and appearance fusion  was done by \cite{Kazakos_2019_ICCV}. %which alone are not enough to adequately adapt to unseen data. \\

\noindent
\textbf{Baseline Methods.} \label{par:baselinemethods} To empirically prove the limitations caused by strictly enforcing an alignment or orthogonality between RGB and audio representations (see Section~\ref{sec:assumptions}), we compare our $\mathcal{L}_{RNA}$ with an \textit{alignment-based} and an \textit{orthogonality-based} loss respectively, both operating on the features of the two modalities. The first, which we indicate with \textit{$\mathcal{L}_{\parallel}$}, imposes an alignment constraint by minimizing the term $1-CosineSimilarity(x,y)$, ideally aiming to the representation in Figure \ref{fig:OurLoss}-b. The second, which we indicate with \textit{$\mathcal{L}_{\bot}$}, operates by minimizing the term $CosineSimilarity(x,y)^2$, imposing an \textit{orthogonality} constraint (Figure \ref{fig:OurLoss}-c). To demonstrate that mitigating the unbalance between the modality feature norms helps the classifier to better exploit the two modalities, we add a Batch Normalization layer before the $G^{m}$ classifier, that serves as a regularizer on input features. We adapt all these baseline methods to our backbone architecture, in order to fairly compare them with our RNA-Net. The baseline for single-source DG is the standard \textit{source-only} approach, while in a multi-source context we take as baseline %We validate our method in a multi-source context against 
the so-called \textit{Deep All} approach, namely
%This indicates the performance of 
the backbone architecture when no other domain adaptive strategies are exploited and \textit{all and only} the source domains are fed to the network. Indeed, this is the ultimate validation protocol in image-based DG methods \cite{bucci2020selfsupervised}. We also provide as a competitor a self-supervised approach, inspired by works that proved its robustness across-domains \cite{carlucci2019domain}. The choice fell on a multi-modal \textit{synchronization} task~\cite{Munro_2020_CVPR}. 

\noindent
\textbf{DG Results.} Table \ref{tab:single-source-dg}-a shows single-source DG results. We see that  $\mathcal{L}_{\bot}$ (referred to as \textit{orthogonality only}) outperforms $\mathcal{L}_{\parallel}$ (referred to as \textit{alignment only}) by up to $1\%$. This confirms that preserving modality-specific features guides the network in the right direction. RNA-Net outperforms such methods by up to $3\%$, confirming that bounding the features in a mutually exclusive aligned or orthogonal space representation could cause a degradation in performance. At the same time, the need of balancing between the two norm distributions is shown to be effective by the results obtained by adding a simple regularization strategy (referred to \textit{BatchNorm}). Once again, our RNA-Net outperforms the competitors by up to $3\%$, proving the strength of $\mathcal{L}_{RNA}$. Finally, the fact that a robust method as the self-supervised (referred as \textit{SS}) does not surpass  the source-only baseline, highlights the complexity of the problem. %while proving our result ($+3.88\%$) to be remarkable. %Moreover, this points out the validity of the baseline choices described in the above paragraph. 
Table \ref{tab:single-source-dg}-b shows the results obtained on  multi-source DG. Our method achieves a consistent boost in performance ($+6.4\%$) w.r.t. DeepAll, and outperforms all other baselines.

\noindent
\textbf{DA Results.} We validate or method in the DA context against four existing unsupervised domain adaptation approaches: 
%\begin{itemize}
 %   \item 
 (a) AdaBN \cite{ada-bn}: Batch Normalization layers are updated with target domain statistics;
 %   \item 
 (b) MMD \cite{da-mmdlong2015learning}: it minimizes separate discrepancy measures applied to single modalities;
 %   \item 
 (c) Adversarial Only \cite{grl-pmlr-v37-ganin15}: a domain discriminator is trained in an adversarial fashion through the gradient reverse layer (GRL) in order to make the feature representations for source and target data indistinguishable;
 %   \item 
 (d) MM-SADA \cite{Munro_2020_CVPR}: a multi-modal domain adaptation framework which is based on the combination of existing DA methods, i.e., a self-supervised synchronization pretext task and an adversarial approach.
%\end{itemize}

Results are summarized in Table \ref{tab:multi-modal-da}. When target data is available at training time, our $\mathcal{L}_{RNA}$ outperforms the standard DA approaches AdaBN \cite{ada-bn} and MMD \cite{da-mmdlong2015learning} by $5.8\%$ and $3.3\%$ respectively. Moreover, our method outperforms adversarial alignment \cite{grl-pmlr-v37-ganin15} by $4\%$. Interestingly, when used in combination to the adversarial approach, our $\mathcal{L}_{RNA}$ slightly improves performances. This complementarity is due to ability of our approach to preserve the structural discrimination of each modality and its intra-class compactness, compensating the distortion in the original distribution induced by the adversarial approach. This validates the considerations done at the end of Section \ref{sec:rna_loss}. Conversely, $\mathcal{L}_{RNA}$  achieves a boost of more than $1\%$ in terms of accuracy when compared against a standard self-supervised synchronization task, which in turn operates by means of reducing the discrepancy, as we do. Finally, we validate our method against the most recent approach in video-based DA literature, i.e., MM-SADA \cite{Munro_2020_CVPR}, achieving on-par results. Considering that MM-SADA combines both a self-supervised and adversarial approach, we compete by means of a lightweight architecture and by employing different modalities.  

\begin{figure}[t]
    \centering
    \includegraphics[width=0.95\linewidth]{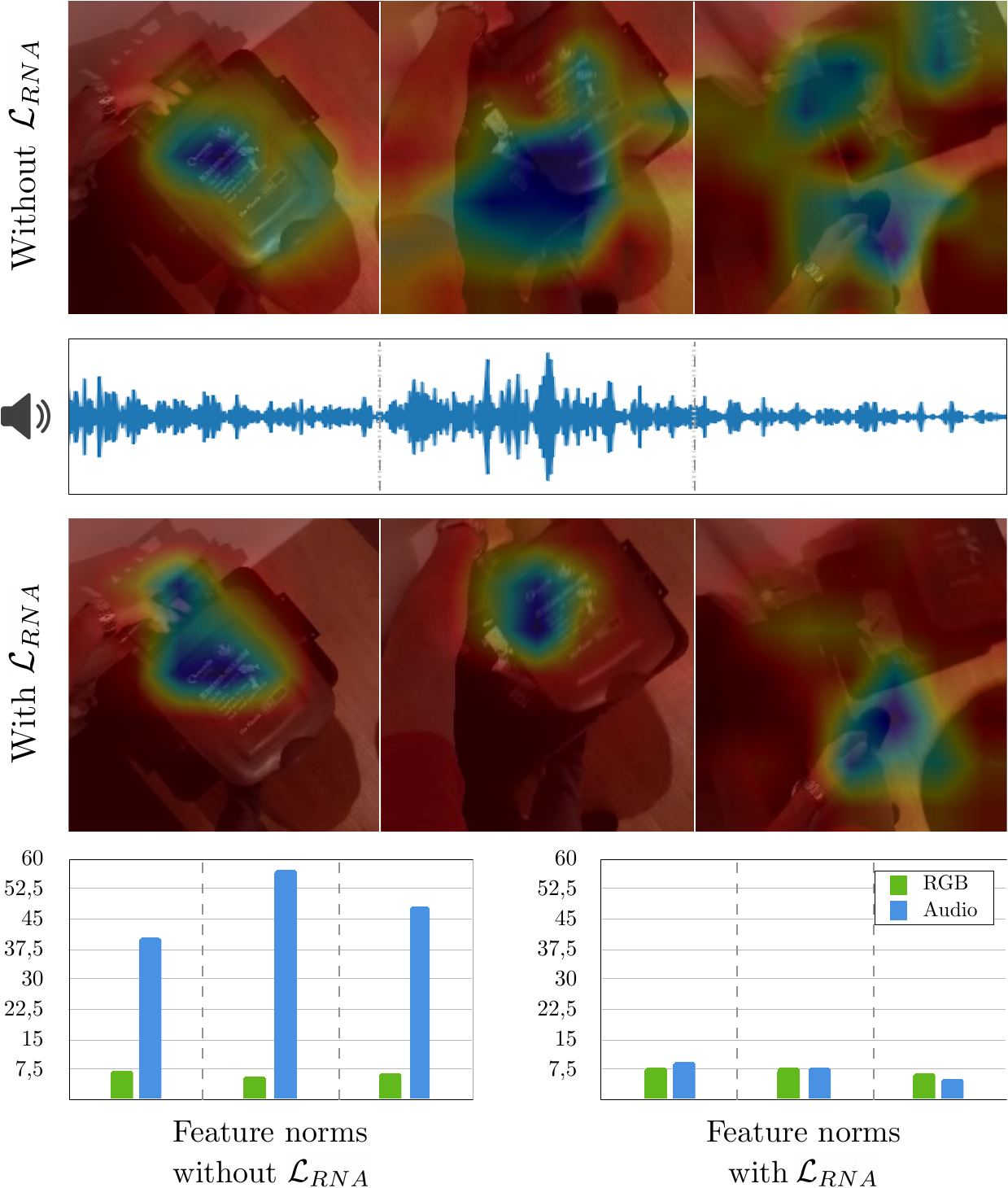}
    \caption{\textbf{Qualitative DG results.} Class activation maps (CAMs) obtained on a target segment using a model trained without (top) and with (bottom) our proposed $\mathcal{L}_{RNA}$ loss, with its audio waveform (middle). A benefit brought by our method is a more localized focus that the network puts on relevant portions of the image after re-balancing the contribution of the two modalities (\textcolor{blue}{blue} corresponds to higher attention, \textcolor{red}{red} to less). The effects on the feature norm values are visible in the histograms at the bottom.
    }
    \label{fig:cams}
    \vspace{-0.1cm}
\end{figure}

\begin{figure}[t]
    \centering
    \includegraphics[width=1\linewidth]{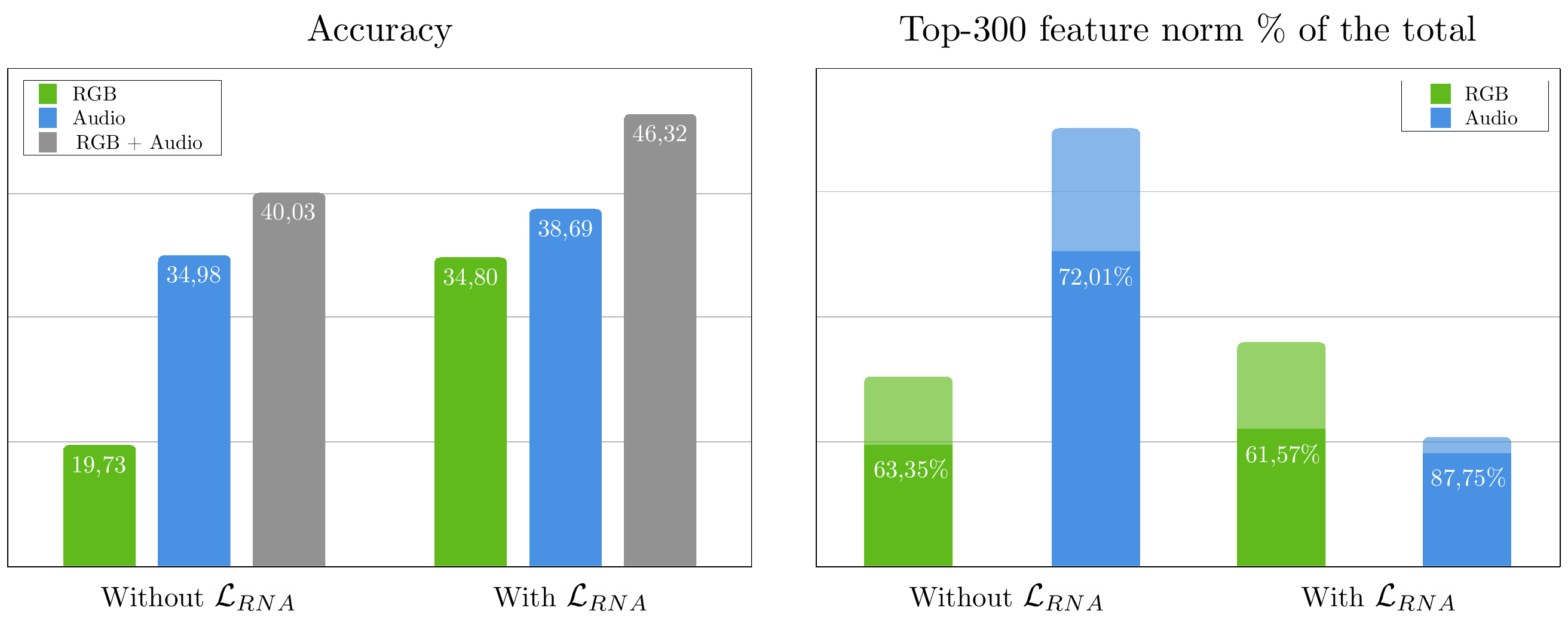}
    \caption{Final score prediction unbalance between audio and visual modalities w/ and w/o our loss function (left). Discrepancy between the norm ranges and their variation before and after the adjustment (right). When minimizing $\mathcal{L}_{RNA}$, the features which are kept active are the most relevant for classification, i.e., top-300.
}
    \label{fig:PlotAccNorm}
    \vspace{-0.3cm}
\end{figure}

\textbf{Ablation Study.} To verify the effectiveness of our design choice, we introduce a loss variant, called \textit{Hard Norm Alignment} (HNA), that induces the norms to tend to a given value arbitrarily $R$. The $R$ term is chosen after observing the range of the norms of the two modalities, and picking a value half-way between the two. To further prove the strength of our method over different architectural variants, we compare the \textit{late fusion} approach against the so-called \textit{mid-level fusion}, proposed in \cite{Kazakos_2019_ICCV}. %It consists in feeding the prediction layer with the fusion of the two modality features, i.e., concatenation.
It consists in feeding the prediction layer a concatenation of the two modality features.
The results are shown in Table \ref{tab:ablation_}. Note how HNA performs worse than $\mathcal{L}_{RNA}$ in all contexts, confirming that an ``hard'' loss function constitutes in a limit. %Instead, letting the proportion between the norms of the two self-reweighting by our $\mathcal{L}_{RNA}$ leads to better results (up to $1\%$). 
As far as concern the mid-level fusion approach, it demonstrates to be a valid alternative in both the supervised and cross-domain settings, remarking the flexibility of our method to be employed in different feature fusion strategies. \\
\input{latex/Tables/Ablation}
\textbf{Qualitative Analysis.} We give an insight of the norm unbalance problem described in Section \ref{sec:assumptions} by showing diagrams representing the norm variations and their impact on the performance. To the readjustment of the norms corresponds a boost in performance (Figure \ref{fig:PlotAccNorm}-a). We also show in Figure \ref{fig:PlotAccNorm}-b the percentage of the total norm given by the 300 most relevant features for classification. While minimizing $\mathcal{L}_{RNA}$, the top 300 features maintain (or even increase) their importance, since their norm ends up representing the majority of the total norm. This further remarks that while relatively adjusting the feature norms of the two modalities, our $\mathcal{L}_{RNA}$ serves as a feature ``selector" for the final classifier.
Lastly, our method brings the side benefit of making the network focus more on relevant portions of the image, with sharper and well defined class activation maps (CAMs) w.r.t. the baseline, as shown in Figure \ref{fig:cams}.

\label{tab:ablation}


%% file: latex/Tables/Ablation.tex
\begin{table}
\centering
\begin{adjustbox}{width=1.0\columnwidth, margin=0ex 1ex 0ex 0ex}
\begin{tabular}{l|cccc}
            \hline \hline
 Ablations          & Supervised     & Single-DG      & Multi-DG       & DA              \\ \hline \hline
Baseline   & 59.76          & 40.93          & 44.67          & -               \\
Fusion     & 60.18          & 40.33          & 47.61          & -               \\
HNA        & 62.41          & 44.58          & 46.70           & 47.19           \\
RNA-Net Fusion & 62.11          & 45.48          & 49.56          & 45.73           \\
\hline
RNA-Net        & \textbf{63.13} & \textbf{45.75} & \textbf{51.06} & \textbf{47.71}  \\
\hline
\end{tabular}
\end{adjustbox}
\caption{Top-1 Accuracy ($\%$) of RNA-Net w.r.t. its \textit{mid-fusion} implementation (RNA-Net Fusion) and HNA on all settings. }
\vspace{-0.39cm}
\label{tab:ablation_}
\end{table}

%% file: latex/Text/Conclusion.tex
\section{Conclusion}

In this work we show the importance of auditory information in  cross-domain first person action recognition. We exploit the complementary nature
of audio and visual information by defining a new cross-modal loss function that operates directly on the relative feature norm of the two modalities. Extensive experiments on DG and DA settings prove the power of our loss. Future work will further pursue this research avenue, exploring the effectiveness of the RNA-loss in third person activity recognition settings, and combined with traditional cross-domain architectures.

%By revealing the importance of exploiting audio-visual complementarity to tackle the domain shift issue, we propose a new cross-modal loss function which directly operates on the relative feature norm of the two modalities. Through extensive experiments, we empirically prove our Relative Norm Alignment loss to be a simple yet effective method to generalize and adapt on unseen data under both the DG and UDA settings.